\documentclass[10pt,a4paper,conference]{IEEEtran}
\usepackage{float}
\usepackage{amsmath}
\usepackage{graphicx}
\usepackage{epstopdf}
\usepackage{amsfonts}
\usepackage{hyperref}
\usepackage{xcolor}
\usepackage{adjustbox}
\usepackage{multirow}
\usepackage{subcaption}
\usepackage{cite}

\begin{document}

\title{Efficient Sentence Embedding via Semantic Subspace Analysis}

\author{\IEEEauthorblockN{Bin Wang}
\IEEEauthorblockA{Univ. of Southern California \\
Los Angeles, USA\\
\texttt{wang699@usc.edu}}
\and
\IEEEauthorblockN{Fenxiao Chen}
\IEEEauthorblockA{Univ. of Southern California \\
Los Angeles, CA, USA\\
\texttt{fenxiaoc@usc.edu}}
\and
\IEEEauthorblockN{Yuncheng Wang}
\IEEEauthorblockA{Univ. of Southern California \\
Los Angeles, CA, USA\\
\texttt{yunchenw@usc.edu}}
\and
\IEEEauthorblockN{C.-C. Jay Kuo}
\IEEEauthorblockA{Univ. of Southern California \\
Los Angeles, CA, USA\\
\texttt{cckuo@sipi.usc.edu}}}

\author{\IEEEauthorblockN{Bin Wang, Fenxiao Chen, Yuncheng Wang, and C.-C. Jay Kuo}
\IEEEauthorblockA{Signal and Image Processing Institute \\
Department of Electrical and Computer Engineering \\
University of Southern California \\
Los Angeles, CA, USA \\
\texttt{\{wang699,fenxiaoc,yunchenw\}@usc.edu},\ \texttt{cckuo@sipi.usc.edu}
}
}

\maketitle

\begin{abstract}
A novel sentence embedding method built upon semantic subspace analysis,
called semantic subspace sentence embedding (S3E), is proposed in this
work. Given the fact that word embeddings can capture semantic
relationship while semantically similar words tend to form semantic
groups in a high-dimensional embedding space, we develop a sentence
representation scheme by analyzing semantic subspaces of its constituent
words. Specifically, we construct a sentence model from two aspects.
First, we represent words that lie in the same semantic group using the
intra-group descriptor. Second, we characterize the interaction between
multiple semantic groups with the inter-group descriptor. The proposed
S3E method is evaluated on both textual similarity tasks and supervised
tasks. Experimental results show that it offers comparable or better performance than the state-of-the-art. The complexity of our S3E method is also much lower than other parameterized models.
\end{abstract}

\IEEEpeerreviewmaketitle

\section{Introduction}\label{sec:introduction}

    Word embedding technique is widely used in natural language
    processing (NLP) tasks. For example, it improves downstream tasks such
    as machine translation \cite{MT}, syntactic parsing \cite{Pars}, and
    text classification \cite{Shen2018Baseline}. Yet, many NLP applications
    operate at the sentence level or a longer piece of texts. Although
    sentence embedding has received a lot of attention recently, encoding a
    sentence into a fixed-length vector to capture different linguistic
    properties remains to be a challenge. 

    Universal sentence embedding aims to compute sentence representation that can be applied to any tasks. It can be categorized into two types:
    i) parameterized models and ii) non-parameterized models.  Parameterized
    models are mainly based on deep neural networks and demand training in their parameter updates. Inspired by the famous word2vec model \cite{mikolov2013distributed}, the
    skip-thought model \cite{kiros2015skip} adopts an encoder-decoder model
    to predict context sentences in an unsupervised manner. InferSent
    model \cite{conneau2017supervised} is trained by high quality supervised
    data; namely, the Natural Language Inference data. It shows that supervised training objective can outperform unsupervised ones. USE \cite{USE} combines both supervised and unsupervised objectives and transformer architecture is employed. The STN
    model \cite{STN} leverages a multi-tasking framework for sentence
    embedding to provide better generalizability.  With the recent success of deep contextualized word models, SBERT \cite{reimers-2019-sentence-bert} and SBERT-WK \cite{SBERT-WK} are proposed to leverage the power of self-supervised learning from large unlabeled corpus. Different parameterized models attempt to capture semantic
    and syntactic meanings from different aspects. Even though their performance is better as compared with non-parameterized models, parameterized ones are more complex and computationally expensive. Since it is challenging to deploy parameterized models into mobile or terminal devices, finding effective and efficient sentence embedding models are necessary.

    \begin{figure}[t]
    \centering
    \includegraphics[width=0.5\textwidth]{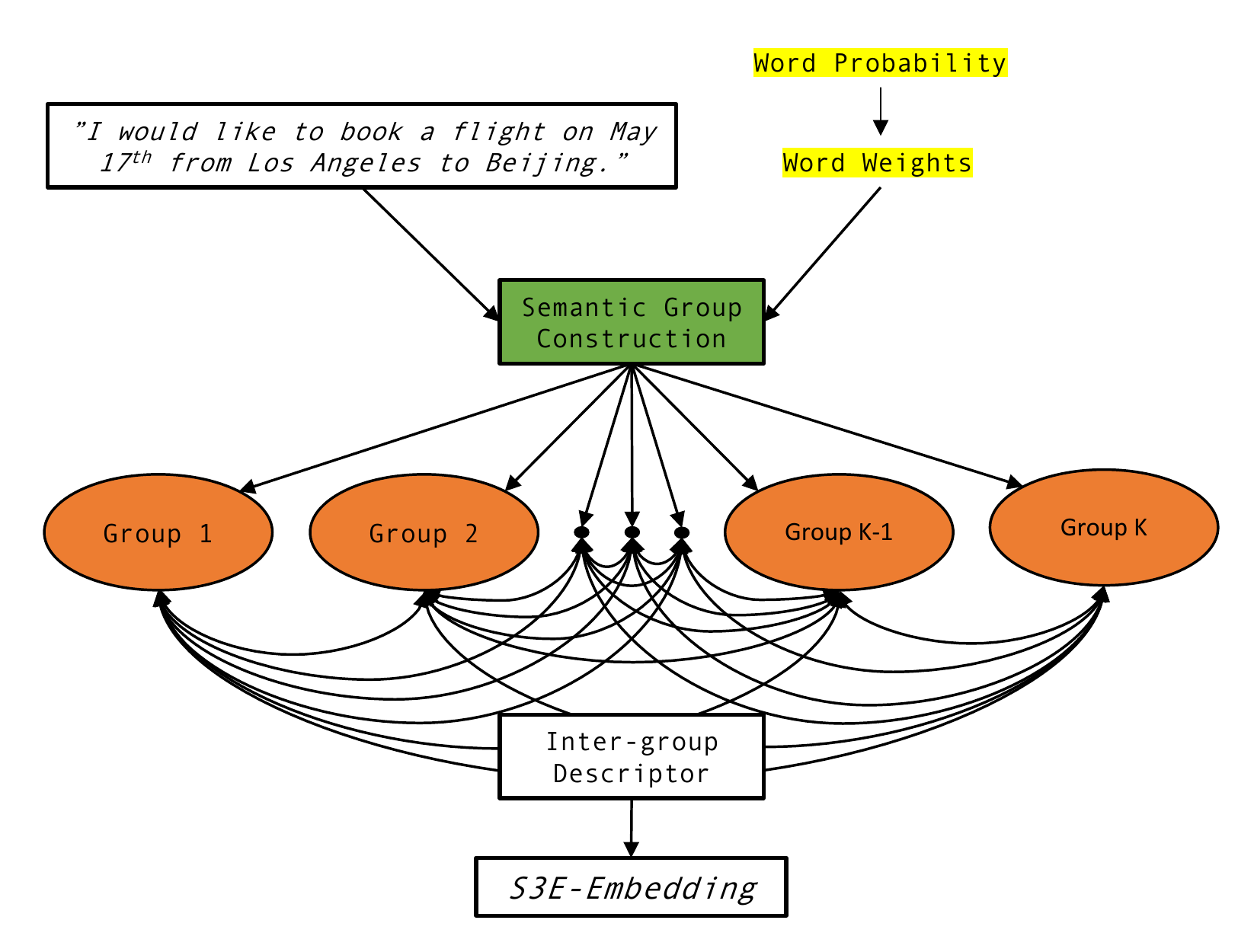}
    \caption{Overview of the proposed S3E method.}\label{S3E}
    \end{figure}

    Non-parameterized sentence embedding methods rely on high quality word
    embeddings. The simplest idea is to average individual word embeddings,
    which already offers a tough-to-beat baseline.  By following along this
    line, several weighted averaging methods have been proposed, including
    tf-idf, SIF \cite{arora2017asimple}, 
    and GEM \cite{GEM}.  Concatenating vector representations of different resources
    yields another family of methods. Examples include SCDV \cite{SCDV} and
    $p$-mean \cite{ruckle2018concatenated}. To better capture the sequential
    information, DCT \cite{DCT} and EigenSent \cite{EigenSent} were proposed
    from a signal processing perspective. 

    Here, we propose a novel non-parameterized sentence embedding method
    based on semantic subspace analysis. It is called semantic subspace
    sentence embedding (S3E) (see Fig. \ref{S3E}). The S3E method is
    motivated by the following observation.  Semantically similar words tend
    to form semantic groups in a high-dimensional embedding space.  Thus, we
    can embed a sentence by analyzing semantic subspaces of its constituent
    words.  Specifically, we use the intra- and inter-group descriptors to
    represent words in the same semantic group and characterize interactions
    between multiple semantic groups, respectively.

    This work has three main contributions. 
    \begin{enumerate} 
    \item The proposed S3E method contains three steps: 1) semantic group
    construction, 2) intra-group descriptor and 3) inter-group descriptor.
    The algorithms inside each step are flexible and, as a result, previous
    work can be easily incorporated. 
    \item To the best of our knowledge, this is the first work that leverages
    correlations between semantic groups to provide a sentence descriptor.
    Previous work using the covariance descriptor \cite{Cov} yields
    super-high embedding dimension (e.g. 45K dimensions). In contrast, the
    S3E method can choose the embedding dimension flexibly. 
    \item The effectiveness of the proposed S3E method in textual similarity
    and supervised tasks is shown experimentally.  Its performance is as
    competitive as that of very complicated parametrized
    models.\footnote{Our code is available at \href{https://github.com/BinWang28/Sentence-Embedding-S3E}{github}.}
    \end{enumerate}

\section{Related Previous Work}
  
    Vector of Locally Aggregated Descriptors (VLAD) is a famous algorithm in the image retrieval field. Same with Bag-of-words method, VLAD trains a codebook based on clustering techniques and concatenate the feature within each clusters as the final representation. Recently work called VLAWE (vector of locally-aggregated word embeddings) \cite{VLAWE}, introduce this idea into document representation. However, VLAWE method suffers from high dimensionality problem which is not favored by machine learning models. In this work, a novel clustering method in proposed by taking word frequency into consideration. At the same time, covariance matrix is used to tackle the dimensionality explosion problem of VLAWE method.

    Recently, a novel document distance metric called Word Mover's Distance (WMD) \cite{kusner2015word} is proposed and achieved good performance in classification tasks. Based on the fact that semantically similar words will have close vector representations, the distance between two sentences are models as the minimal 'travel' cost for moving the embedded words from one sentence to another. WMD targets on modeling the distance between sentences in the shared word embedding space. It is natural to consider the possibility of computing the sentence representation directly from the word embedding space by semantical distance measures.

    There are a few works trying to obtain sentence/document representation based on Word Mover's Distance. D2KE (distances to kernels and embeddings) and WME (word mover's embedding) converts the distance measure into positive definite kernels and has better theoretical guarantees. However, both methods are proposed under the assumption that Word Mover's Distance is a good standard for sentence representation. In our work, we borrow the 'travel' concept of embedded words in WMD's method. And use covariance matrix to model the interaction between semantic concepts in a discrete way.

\section{Proposed S3E Method}\label{sec:approach}
  
    As illustrated in Fig. \ref{S3E}, the S3E method contains three steps:
    1) constructing semantic groups based on word vectors; 2) using the
    inter-group descriptor to find the subspace representation; and 3) using
    correlations between semantic groups to yield the covariance descriptor.
    Those are detailed below. 

    {\bf Semantic Group Construction.} Given word $w$ in the vocabulary, $V$,
    its uni-gram probability and vector are represented by $p(w)$ and $v_w
    \in \mathbb{R}^d$, respectively. We assign weights to words based on $p(w)$:
    \begin{equation}\label{eq:epsilon}
    \mbox{weight}(w) = \frac{\epsilon}{\epsilon+p(w)},
    \end{equation}
    where $\epsilon$ is a small pre-selected parameter, which is added to
    avoid the explosion of the weight when $p(w)$ is too small. Clearly, $0
    < \mbox{weight}(w) < 1$.  Words are clustered into $K$ groups using the
    K-means++ algorithm \cite{k-means++}, and weights are incorporated in
    the clustering process. This is needed since some words of higher
    frequencies (e.g.  \textit{'a', 'and', 'the'}) are less discriminative
    by nature. They should be assigned with lower weights in the semantic
    group construction process. 

    {\bf Intra-group Descriptor.} After constructing semantic groups, we
    find the centroid of each group by computing the weighted average of
    word vectors in that group. That is, for the $i_{th}$ group, $G_i$, we
    learn its representation $g_i$ by
    \begin{equation}
    g_i = \frac{1}{|G_i|}\sum_{w\in G_i} \mbox{weight}(w) v_w,
    \end{equation}
    where $|G_i|$ is the number of words in group $G_i$.  For sentence $S =
    \{w_1, w_2, ..., w_m\}$, we allocate words in $S$ to their semantic
    groups. To obtain the intra-group descriptor, we compute the cumulative
    residual between word vectors and their centroid ($g_i$) in the same group. 
    Then, the representation of sentence $S$ in the $i_{th}$ semantic group 
    can be written as
    \begin{equation}
    v_i = \sum_{w\in S\cap G_i} \mbox{weight}(w)(v_w - g_i).
    \end{equation}
    If there are $K$ semantic groups in total, we can represent sentence $S$ 
    with the following matrix:
    \begin{equation}\label{eq:phi}
    \Phi(S) = 
      \begin{bmatrix}
        v_{1}^T  \\
        v_{2}^T  \\
        \vdots   \\
        v_{K}^T 
        \end{bmatrix}
        =
        \begin{pmatrix}
        v_{11} & \dots  & v_{1d} \\
        v_{21} & \dots  & v_{2d} \\
        \vdots & \ddots & \vdots \\
        v_{K1} & \dots  & v_{Kd}
      \end{pmatrix}_{K\times d},
    \end{equation}
    where $d$ is the dimension of word embedding.

    {\bf Inter-group Descriptor.} After obtaining the intra-group
    descriptor, we measure interactions between semantic groups with
    covariance coefficients.  We can interpret $\Phi(S)$ in (\ref{eq:phi}),
    as $d$ observations of $K$-dimensional random variables, and use
    $u_\Phi\in \mathbb{R}^{K\times 1}$ to denote the mean of each row 
    in $\Phi$. Then, the inter-group covariance matrix can be computed as
    \begin{equation}
    C = [ C_{i,j} ]_{K \times K} = \frac{1}{d}(\Phi-\mu_\Phi)(\Phi-\mu_\Phi)^T \in \mathbb{R}^{K \times K},
    \end{equation}
    where 
    \begin{equation}
    C_{i,j} = \sigma_{i,j}=\frac{(v_i-\mu_i)^T(v_j-\mu_j)}{d}.
    \end{equation}
    is the covariance between groups $i$ and $j$. Thus, matrix $C$ can be written as
    \begin{equation}
    C = 
    \begin{pmatrix}
    \sigma_{1}^2 & \sigma_{12}  & \dots  & \sigma_{1K} \\
    \sigma_{12}  & \sigma_{2}^2 & \dots  & \sigma_{2K} \\
    \vdots       & \vdots       & \ddots & \vdots      \\
    \sigma_{1K}  & \sigma_{2K}  & \dots  & \sigma_{K}^2
    \end{pmatrix}.
    \end{equation}
    Since the covariance matrix $C$ is symmetric, we can vectorize its upper
    triangular part and use it as the representation for sentence $S$. The Frobenius norm of the original matrix is kept the same with the Euclidean norm of vectorized matrices. This process produces an embedding of dimension $\mathbb{R}^{K(K+1)/2}$. Then,
    the embedding of sentence $S$ becomes
    \begin{equation}\label{eq:sentence_embedding}
    v(S) = vect(C) = 
    \left\{
    \begin{array}{ll}
    \sqrt{2}\sigma_{ij}, & \text{if}\quad i<j, \\
    \sigma_{ii},         & \text{if}\quad i=j.
    \end{array}
    \right.
    \end{equation}
    Finally, the sentence embedding in Eq. (\ref{eq:sentence_embedding}) is
    $L_2-\text{normalized}$.

    {\bf Complexity}

        The semantic group construction process can be pre-computed for
        efficiency. Our runtime complexity is ($dN+Kd^2$), where $N$ is the length
        of a sentence, $K$ is the number of semantic groups, and $d$ is the
        dimension of word embedding in use. Our algorithm is linear with respect to the sentence length. The S3E method is much faster than
        all parameterized models and most of non-parameterized methods such as
        \cite{GEM} where the singular value decomposition is needed during
        inference. The run time comparison is also discussed in Sec. \ref{sec:runtime}.

\section{Experiments}\label{sec:experiments}

    We evaluate our method on two sentence embedding evaluation tasks to verify the generalizability of S3E. Semantic textual similarity tasks are used to test the clustering and retrieval property of our sentence embedding. Discriminative power of sentence embedding is evaluated by supervised tasks.

    For performance benchmarking, we compare S3E with a series of other methods including parameterized and non-parameterized ones.

    \begin{enumerate}
        \item Non-parameterized Models
        \begin{enumerate}
             \item[a)] Avg. GloVe embedding;
             \item[b)] SIF \cite{arora2017asimple}: Derived from an improved random-walk model. Consist of two parts: weighted averaging of word vectors and first principal component removal;
             \item[c)] $p$-means \cite{ruckle2018concatenated}: Concatenating different word embedding models and different power ratios;
             \item[d)] DCT \cite{DCT}: Introduce discrete cosine transform into sentence sequential modeling;
             \item[e)] VLAWE \cite{VLAWE}: Introduce VLAD (vector of locally aggregated descriptor) into sentence embedding field;
         \end{enumerate}
        \item Parameterized Models
        \begin{enumerate}
             \item[a)] Skip-thought \cite{kiros2015skip}: Extend word2vec unsupervised training objectives from word level into sentence level;
             \item[b)] InferSent \cite{conneau2017supervised}: Bi-directional LSTM encoder trained on high quality sentence inference data.
             \item[c)] Sent2Vec \cite{sent2vec}: Learn n-gram word representation and use average as the sentence representation.
             \item[d)] FastSent \cite{fastsent}: An improved Skip-thought model for fast training on large corpus. Simplify the recurrent neural network as bag-of-words representation.
             \item[e)] ELMo \cite{ELMO}: Deep contextualized word embedding. Sentence embedding is computed by averaging all LSTM outputs.
             \item[f)] Avg. BERT embedding \cite{devlin2018bert}: Average the last layer word representation of BERT model.
             \item[g)] SBERT-WK \cite{SBERT-WK}: A fusion method to combine representations across layers of deep contextualized word models.
        \end{enumerate}
    \end{enumerate}

    \subsection{Textual Similarity Tasks}

        \begin{table*}[htb]
        \centering
        \caption{Experimental results on textual similarity tasks in terms of the Pearson 
        correlation coefficients (\%), where the best results for parameterized and non-parameterized are in bold respectively.}
        \resizebox{1.8\columnwidth}{!}{
        \begin{tabular}{ c | c | c | c | c | c | c | c | c || c } 
        \hline
        Model & Dim & STS12 & STS13 & STS14 & STS15 & STS16 & STSB & SICK-R & Avg.\\ 
        \hline\hline
        \multicolumn{3}{l}{\texttt{Parameterized models}} \\
        \hline\hline
        skip-thought\cite{kiros2015skip} & 4800 & 30.8 & 24.8 & 31.4 & 31.0 & - & - & 86.0 & 40.80 \\
        InferSent\cite{conneau2017supervised} & 4096 & 58.6 & 51.5 & 67.8 & 68.3 & 70.4 & 74.7 & \textbf{88.3} & 68.51\\
        ELMo\cite{ELMO} & 3072 & 55.0 & 51.0 & 63.0 & 69.0 & 64.0 & 65.0 & 84.0 & 64.43\\ 
        Avg. BERT \cite{devlin2018bert} & 768 & 46.9 & 52.8 & 57.2 & 63.5 & 64.5 & 65.2 & 80.5 & 61.51 \\
        SBERT-WK \cite{SBERT-WK} & 768 & \textbf{70.2} & \textbf{68.1} & \textbf{75.5} & \textbf{76.9} & \textbf{74.5} & \textbf{80.0} & 87.4 & \textbf{76.09} \\
        \hline\hline
        \multicolumn{3}{l}{\texttt{Non-parameterized models}} \\
        \hline\hline
        Avg. GloVe & 300 & 52.3 & 50.5 & 55.2 & 56.7 & 54.9 & 65.8 & 80.0 & 59.34 \\
        SIF\cite{arora2017asimple} & 300 & 56.2 & 56.6 & 68.5 & 71.7 & - & 72.0 & \textbf{86.0} & 68.50 \\
        $p$-mean\cite{ruckle2018concatenated} & 3600 & 54.0 & 52.0 & 63.0 & 66.0 & 67.0 & 72.0 & \textbf{86.0} & 65.71 \\
        \hline\hline
        S3E (GloVe)  & 355-1575 & 59.5 & 62.4 & 68.5 & 72.3 & 70.9 & 75.5 & 82.7 & 69.59 \\
        S3E (FastText) & 355-1575 & \textbf{62.5} & 67.8 & 70.2 & \textbf{76.1} & 74.3 & 77.5 & 84.7 & 72.64 \\
        S3E (L.F.P.) & 955-2175 & 61.0 & \textbf{69.3} & \textbf{73.2} & \textbf{76.1} & \textbf{74.4} & \textbf{78.6} & 84.7 & \textbf{73.90}\\ \hline
        \end{tabular}}
        \label{STS}
        \end{table*}

        We evaluate the performance of the S3E method on the SemEval semantic
        textual similarity tasks from 2012 to 2016, the STS Benchmark and SICK-Relatedness dataset. The goal
        is to predict the similarity between sentence pairs. The sentence pairs contains labels between 0 to 5, which indicate their semantic relatedness. The Pearson
        correlation coefficients between prediction and human-labeled
        similarities are reported as the performance measure. For STS 2012 to 2016 datasets, the similarity
        prediction is computed using the cosine similarity. For STS Benchmark dataset and SICK-R dataset, they are under supervised setting and aims to predict the probability distribution of relatedness scores. We adopt the same setting with \cite{tai2015improved} for these two datasets and also report the Pearson correlation coefficient.

        The S3E method can be applied to any static word embedding method. Here, we
        report three of them; namely, GloVe \cite{GloVe}, FastText
        \cite{FastText} and L.F.P. \footnote{concatenated LexVec, FastText and PSL}.  Word embedding is normalized using \cite{PVN}. Parameter $\epsilon$ in Eq.  (\ref{eq:epsilon})
        is set to $10^{-3}$ for all experiments. The word frequency, $p(w)$, is estimated from the wiki
        dataset\footnote{https://dumps.wikimedia.org/}. The number of semantic
        groups, $K$, is chosen from the set $\{ 10,20,30,40,50 \}$ and the 
        best performance is reported.

        Experimental results on textual similarity tasks are shown in Table
        \ref{STS}, where both non-parameterized and parameterized models are
        compared. Recent parameterized method SBERT-WK provides the best performance and outperforms other method by a large margin. S3E method using L.F.P word embedding is the second best method in average comparing with both parameterized and non-parameterized methods. As mentioned, our work is compatible with any
        weight-based methods. With better weighting schemes, the S3E method has
        a potential to perform even better. As choice of word embedding,
        L.F.P performs better than FastText and FastText is better than GloVe vector in Table \ref{STS}, which is
        consistent with the previous findings \cite{wang2019evaluating}. Therefore, choosing more powerful word embedding models can be helpful in performance boost.

    \subsection{Supervised Tasks}

        \begin{table*}[htb]
        \centering
        \caption{Experimental results on supervised tasks, where sentence
        embeddings are fixed during the training process and the best results
        for parameterized and non-parameterized models are marked in bold respectively.}
        \resizebox{1.9\columnwidth}{!}{
        \begin{tabular}{c | c | c | c | c | c | c | c | c | c || c} \hline
        Model & Dim & MR & CR & SUBJ & MPQA & SST & TREC & MRPC & SICK-E & Avg.\\ \hline\hline
        \multicolumn{3}{l}{\texttt{Parameterized models}} \\ \hline\hline
        skip-thought\cite{kiros2015skip} & 4800 & 76.6 & 81.0 & 93.3 & 87.1 & 81.8 & 91.0 & 73.2 & 84.3 & 83.54 \\
        FastSent\cite{fastsent} & 300 & 70.8 & 78.4 & 88.7 & 80.6 & - & 76.8 & 72.2 & - & 77.92 \\
        InferSent\cite{conneau2017supervised} & 4096 & 79.3 & 85.5 & 92.3 & 90.0 & 83.2 & 87.6 & 75.5 & 85.1 & 84.81 \\
        Sent2Vec\cite{sent2vec} & 700 & 75.8 & 80.3 & 91.1 & 85.9 & - & 86.4 & 72.5 & - & 82.00 \\
        USE\cite{USE} & 512 & 80.2 & 86.0 & 93.7 & 87.0 & 86.1 & \textbf{93.8} & 72.3 & 83.3 & 85.30 \\
        ELMo\cite{ELMO} & 3072 & 80.9 & 84.0 & 94.6 & 91.0 & 86.7 & 93.6 & 72.9 & 82.4 & 85.76 \\
        SBERT-WK \cite{SBERT-WK} & 768 & \textbf{83.0} & \textbf{89.1} & \textbf{95.2} & \textbf{90.6} & \textbf{89.2} & 93.2 & \textbf{77.4} & \textbf{85.5} & \textbf{87.90} \\
        \hline\hline 
        \multicolumn{3}{l}{\texttt{Non-parameterized models}} \\ \hline\hline
        GloVe(Ave) & 300 & 77.6 & 78.5 & 91.5 & 87.9 & 79.8 & 83.6 & 72.1 & 79.0 & 81.25 \\ 
        SIF\cite{arora2017asimple} & 300 & 77.3 & 78.6 & 90.5 & 87.0 & 82.2 & 78.0 & - & \textbf{84.6} & 82.60 \\
        p-mean\cite{ruckle2018concatenated} & 3600 & 78.3 & 80.8 & 92.6 & 89.1 & \textbf{84.0} & 88.4 & 73.2 & 83.5 & 83.74 \\
        DCT\cite{DCT} & 300-1800 & 78.5 & 80.1 & 92.8 & 88.4 & 83.7 & \textbf{89.8} & 75.0 & 80.6 & 83.61 \\
        VLAWE\cite{VLAWE} & 3000 & 77.7 & 79.2 & 91.7 & 88.1 & 80.8 & 87.0 & 72.8 & 81.2 & 82.31 \\
        \hline\hline
        S3E (GloVe) & 355-1575 & 78.3 & 80.4 & 92.5 & \textbf{89.4} & 82.0 & 88.2 & 74.9 & 82.0 & 83.46\\
        S3E (FastText) & 355-1575 & 78.8 & \textbf{81.4} & 92.9 & 88.5 & 83.5 & 87.0 & \textbf{75.7} & 81.4 & 83.65 \\
        S3E(L.F.P.) & 955-2175 & \textbf{79.4} & \textbf{81.4} & \textbf{92.9} & \textbf{89.4} & 83.5 & 89.0 & 75.6 & 82.6 & \textbf{84.23}\\
        \hline
        \end{tabular}}
        \label{Supervisedtable}
        \end{table*}

        The SentEval
        toolkit\footnote{https://github.com/facebookresearch/SentEval}
        \cite{conneau2018senteval} is used to evaluate on eight supervised tasks:

        \begin{enumerate}
            \item MR: Sentiment classification on movie reviews.
            \item CR: Sentiment classification on product reviews.
            \item SUBJ: Subjectivity/objective classification.
            \item MPQA: Opinion polarity classification.
            \item SST2: Stanford sentiment treebank for sentiment classification.
            \item TREC: Question type classification.
            \item MRPC: Paraphrase identification.
            \item SICK-Entailment: Entailment classification on SICK dataset.
        \end{enumerate}

        \begin{table}[htb]
            \centering
            \caption{Examples in downstream tasks}\label{table:downstream}
            \resizebox{\columnwidth}{!}{
            \begin{tabular}{| c | c | c | c | c |} 
                \hline
                \textbf{Dataset} & \textbf{\# Samples} & \textbf{Task} & \textbf{Class} \\ \hline
                MR & 11k & movie review & 2  \\\hline
                CR & 4k & product review & 2  \\\hline
                SUBJ & 10k & subjectivity/objectivity & 2 \\\hline
                MPQA & 11k & opinion polarity & 2  \\\hline
                SST2 & 70k & sentiment & 2 \\\hline
                TREC & 6k & question-type & 6\\\hline
                MRPC & 5.7k & paraphrase detection & 2 \\\hline
                SICK-E & 10k & entailment & 3 \\
                \hline
            \end{tabular}}
        \end{table}

        The details for each dataset is also shown in Table \ref{table:downstream}. For all tasks, we trained a simple MLP classifier that contain one hidden layer of 50 neurons. It is same as it was done in \cite{DCT} and only
        tuned the $L_2$ regularization term on validation sets. The
        hyper-parameter setting of S3E is kept the same as that in textual similarity tasks. The batch size is set to 64 and Adam optimizer is employed. For MR, CR, SUBJ, MPQA and MRPC datasets, we use the nested 10-fold cross validation. For TREC and SICK-E, we use the cross validation. For SST2 the standard validation is utilized. All experiments are trained with 4 epochs.

        Experimental results on supervised tasks are shown in Table
        \ref{Supervisedtable}. The S3E method outperforms all non-parameterized
        models, including DCT \cite{DCT}, VLAWE \cite{VLAWE} and $p$-means
        \cite{ruckle2018concatenated}. The S3E method adopts a word embedding
        dimension smaller than $p$-means and VLAWE and also flexible in choosing embedding dimensions. As implemented in other
        weight-based methods, the S3E method does not consider the order of
        words in a sentence but splits a sentence into different semantic
        groups. The S3E method performs the best on the paraphrase
        identification (MRPC) dataset among all non-parameterized and
        parameterized methods excluding SBERT-WK. This is attributed to that, when paraphrasing,
        the order is not important since words are usually swapped. In this
        context, the correlation between semantic components play an important
        role in determining the similarity between a pair of sentences and
        paraphrases.

        Comparing with parameterized method, S3E also outperforms a series of them including Skip-thought, FastSent and Sent2Vec. In general, parameterized methods performs better than non-parameterized ones on downstream tasks. The best performance is the recently proposed SBERT-WK method which incorporate a pre-trained deep contextualized word model. However, even though good perform is witnessed, deep models are requiring much more computational resources which makes it hard to integrate into mobile or terminal devices. Therefore, S3E method has its own strength in its efficiency and good performance.

    \subsection{Inference Speed}
    \label{sec:runtime}
  
        \begin{table}[htb]
            \centering
            \caption{Inference time comparison. Data are collected from 5 trails.}\label{exp:speed}
            \resizebox{\columnwidth}{!}{
                \begin{tabular}{ c | c | c } 
                    \hline
                    Model & CPU inference time (ms) & GPU inference time (ms) \\ \hline
                    InferSent & 53.07  & 15.23 \\ \hline
                    SBERT-WK & 179.27 & 42.79 \\ \hline
                    GEM & 26.54 & - \\ \hline
                    SIF & 1.56 & - \\ \hline
                    Proposed S3E & \textbf{0.69} & - \\ 
                    \hline
                \end{tabular}}
            \end{table}
        We compare the inference speed of S3E with other models including the non-parameterized and parameterized ones. For fair comparison, the batch size is set to 1 and all sentences from STSB datasets are used for evaluation (17256 sentences). All benchmark results can run on CPU\footnote{Intel i7-5930 of 3.50GHz with 12 cores} and GPU\footnote{Nvidia GeForce GTX TITAN X}. The results are showed in Table \ref{exp:speed}.

        Comparing the other method, S3E is the very efficient in inference speed and this is very important in sentence embedding. Without the acceleration of powerful GPU, when doing comparing tasks of 10,000 sentence pairs, deep contextualized models takes about 1 hour to accomplish, which S3E only requires 13 seconds.

    \subsection{Sensitivity to Cluster Numbers}

        \begin{figure}[htb]
            \centering
            \includegraphics[width=\columnwidth]{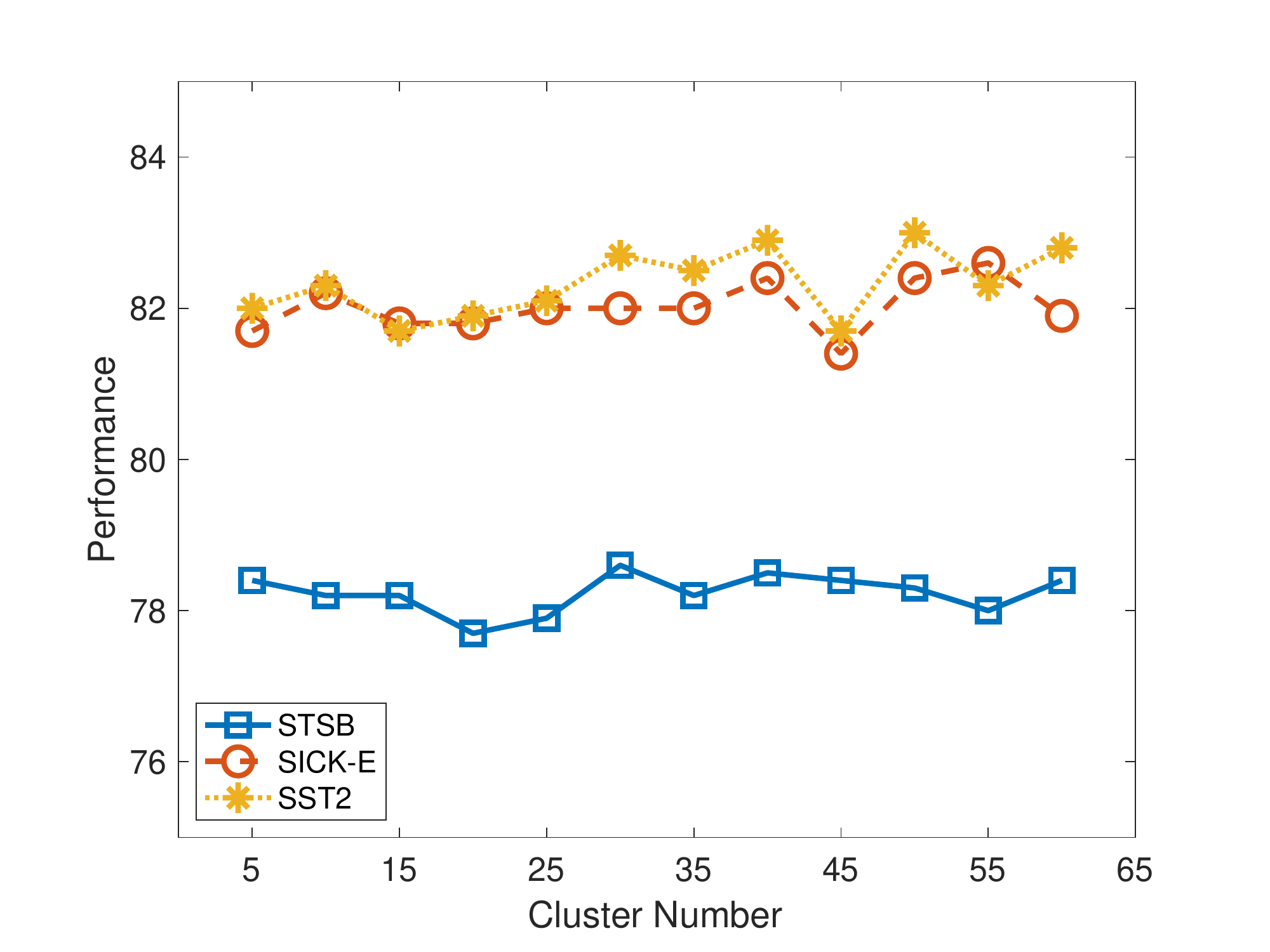}
            \caption{Comparing results with different settings on cluster numbers. STSB result is presented in Pearson Correlation Coefficients (\%). SICK-E and SST2 are presented in accuracy.}
        \label{exp:sensitivity}
        \end{figure}

        We test the sensitivity of S3E to the setting of cluster numbers. The cluster number is set from 5 to 60 with internal of 5 clusters. Results for STS-Benchmark, SICK-Entailment and SST2 dataset are reported. As we can see from Figure \ref{exp:sensitivity}, performance of S3E is quite robust for different choice of cluster numbers. The performance varies less than 1\% in accuracy or correlation.

\section{Discussion}

    Averaging word embedding provides a simple baseline for sentence
    embedding. A weighted sum of word embeddings should offer improvement
    intuitively. Some methods tries to improve averaging are to concatenate word embedding in several forms
    such as $p$-means \cite{ruckle2018concatenated} and VLAWE\cite{VLAWE}.
    Concatenating word embeddings usually encounters the dimension explosion
    problem. The number of concatenated components cannot be too large.

    Our S3E method is compatible with exiting models and its performance can
    be further improved by replacing each module with a stronger one.
    First, we use word weights in constructing semantically similar groups
    and can incorporate different weighting scheme in our model, such as SIF
    \cite{arora2017asimple}, GEM \cite{GEM}. Second,
    different clustering schemes such as the Gaussian mixture model and
    dictionary learning can be utilized to construct semantically similar
    groups \cite{SCDV,guptap}. Finally, the intra-group descriptor can be replaced
    by methods like VLAWE \cite{VLAWE} and $p$-means \cite{ruckle2018concatenated}. In inter-group descriptor, correlation between semantic groups can also be modeled in a non-linear way by applying different kernel functions. Another future direction is to add sequential information into current S3E method.

\section{Conclusion}

    A sentence embedding method based on semantic subspace analysis was
    proposed. The proposed S3E method has three building modules: semantic
    group construction, intra-group description and inter-group description.
    The S3E method can be integrated with many other existing models.  It
    was shown by experimental results that the proposed S3E method offers
    state-of-the-art performance among non-parameterized models. S3E is outstanding 
    for its effectiveness with low computational complexity.

\bibliographystyle{IEEEtran}
\bibliography{IEEEabrv,mybibfile}

\end{document}